\documentclass[letterpaper, 10pt, conference]{ieeeconf/ieeeconf}  
\usepackage[T1]{fontenc}
\usepackage{wrapfig}
\usepackage{booktabs}
\usepackage{amsmath}
\usepackage{amsmath} 
\usepackage{amssymb}  
\usepackage[]{changes}
\definechangesauthor[name={Todor}, color=red]{tsv}
\definechangesauthor[name={Johannes}, color=cyan]{jsk}
\definechangesauthor[name={Martin}, color=orange]{mnmn}
\presetkeys%
{todonotes}%
{inline,backgroundcolor=yellow}{}
\usepackage{chngcntr}
\usepackage{siunitx}
\usepackage{tikz}
\usepackage{adjustbox}
\usepackage{url}
\usepackage{subfigure}
\usepackage{todonotes}
\usepackage{multirow}

\IEEEoverridecommandlockouts                              

\title{\LARGE \bf
 Learning Extrinsic Dexterity with \\Parameterized Manipulation Primitives
}

\author{Shih-Min Yang, Martin Magnusson, Johannes A. Stork, Todor Stoyanov
	\thanks{The authors are with the Center for Applied Autonomous Sensor Systems (AASS), Örebro University, Sweden. 
    Corresponding author {shih-min.yang@oru.se}.}%
  \thanks{*This work has received funding from the EU’s Horizon 2020 research and innovation programme under grant agreement No 101017274, and was supported by the Wallenberg AI, Autonomous Systems and Software Program (WASP) funded by the Knut and Alice Wallenberg Foundation.}
}

\begin{document}
\maketitle
\thispagestyle{empty}
\pagestyle{empty}
\begin{abstract}

    Many practically relevant robot grasping problems feature a target object for which all grasps are occluded, e.g., by the environment.
    Single-shot grasp planning invariably fails in such scenarios. Instead, it is necessary to first manipulate the object into a configuration that affords a grasp.
    %
    We solve this problem by learning a sequence of actions that utilize the environment to change the object's pose.
    %
    Concretely, we employ hierarchical reinforcement learning to combine a sequence of learned parameterized manipulation primitives. 
    %
    By learning the low-level manipulation policies, our approach can control the object's state through exploiting interactions between the object, the gripper, and the environment. 
    Designing such a complex behavior analytically would be infeasible under uncontrolled conditions, as an analytic approach requires accurate physical modeling of the interaction and contact dynamics. In contrast, we learn a hierarchical policy model that operates directly on depth perception data, without the need for object detection, pose estimation, or manual design of controllers. 
    %
    We evaluate our approach on picking box-shaped objects of various weight, shape, and friction properties from a constrained table-top workspace.  
    Our method transfers to a real robot and is able to successfully complete the object picking task in 98\% of experimental trials.
    
\end{abstract}

\section{\uppercase{Introduction}}
\label{sec:introduction}


State-of-the-art robotic grasping systems~\cite{hoang2022context,mahler2018dex,fang2020graspnet} function well in moderately cluttered scenes, but are fundamentally limited in assuming that objects are directly graspable~---~that is, that there always exists a collision-free grasp configuration within the reachability space of the robot arm. 
In practice, this assumption is often violated: for example, in cases when objects are tightly packed together, or placed in configurations that obstruct all feasible grasps (e.g., think of a book lying flat on a table). To address such practically relevant scenarios the robot arm needs to re-arrange objects in a non-prehensile manner, which poses unique challenges to perception, planning, and control.

%
Current non-prehensile object re-arrangement approaches aim to overcome the stochastic and unpredictable nature of physical interaction through trial-and-error learning~\cite{yuan2019end, zhou2022ungraspable, kimpre, zhou2023learning}. 
As reinforcement learning (RL) involving contact dynamics has prohibitively high interaction sample complexity, a common solution is to employ manually designed parametric controllers dubbed \textit{manipulation primitives}~\cite{zeng2018learning, ren2021fast, dalal2021accelerating, nasiriany2022augmenting}. 
However, this has two disadvantages compared to the more general end-to-end approaches: first, it limits applicability to only tasks that can be solved by combining the available primitives; and second, it necessitates expert input in designing, implementing, and tuning the primitive controllers. 
While some progress has recently been made in alleviating the first shortcoming through e.g., the use of atomic actions to ``stitch'' together primitives~\cite{nasiriany2022augmenting}, the need for expert input in primitive design still poses a major challenge.

\begin{figure}[t!]
    \centering
    \includegraphics[width=0.85\linewidth]{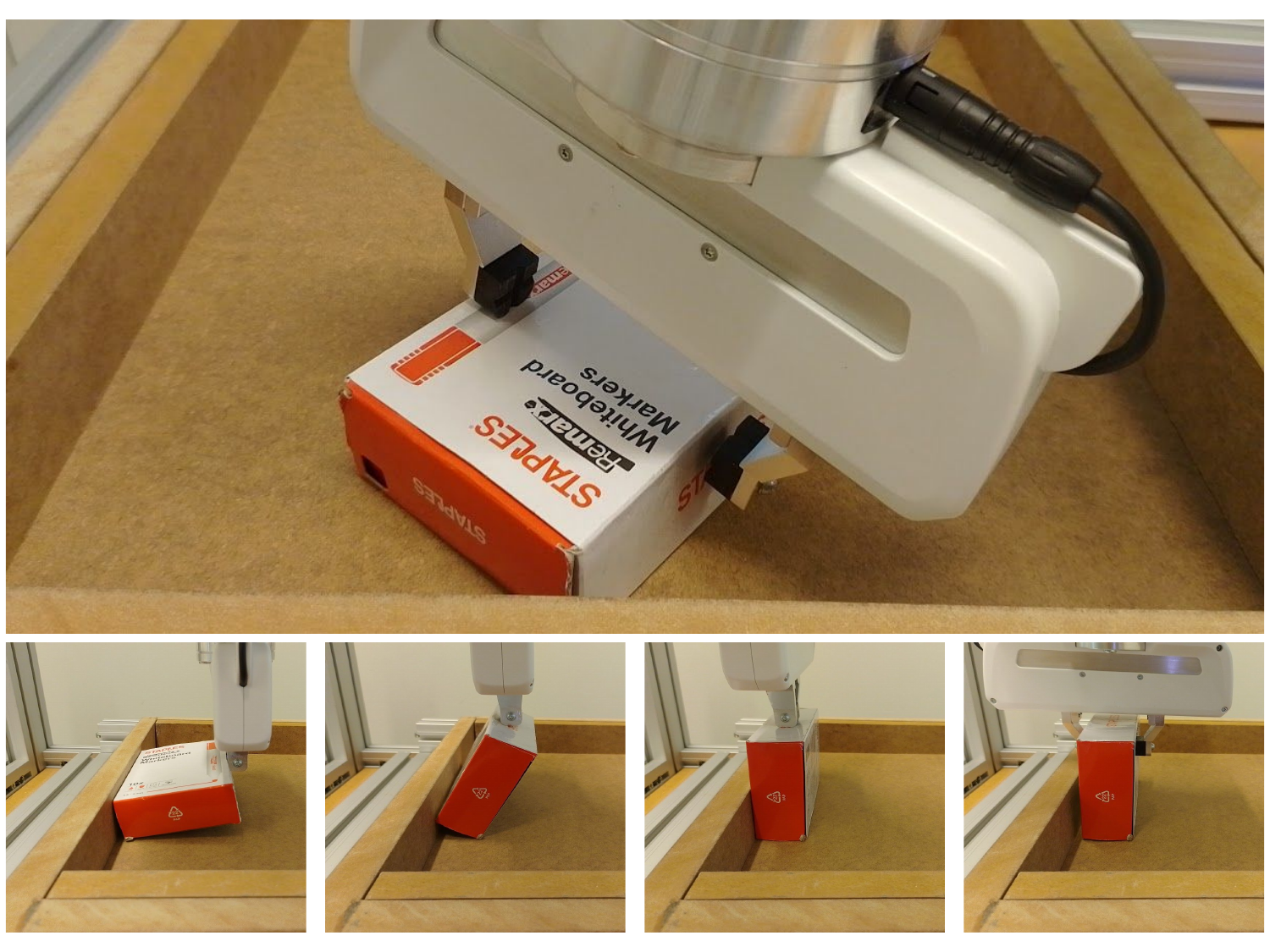}
    \vspace{-0.2cm}
    \caption{
    \emph{Top:} In the initial pose, all feasible grasps on the target object are occluded by the environment.
    \emph{Bottom (left to right):} We learn to push the object to a wall and exploit it as a pivot to flip the object up and finally grasp it from the top.}
    \label{fig: task}
    \vspace{-0.6cm}
\end{figure}

Instead of relying solely on manually-defined primitives or resorting to costly end-to-end RL, we take a middle ground and propose to learn hierarchical control policies whose actions are a series of parametrized learned manipulation primitives. 
This allows us to maintain the generality of full-scale RL while improving learning efficiency through the decomposition of tasks into several primitives, each associated with lower-dimensional state-action spaces.
We apply our approach to solve a variation of the \textit{occluded grasping} task~\cite{zhou2022ungraspable} in which a robot arm equipped with a simple parallel jaw gripper needs to pick a flat object placed on a table-top (see Fig.~\ref{fig: task}). 
As the object is only graspable along approach directions that collide with the table, the task can only be solved via non-prehensile manipulation and interaction with the environment: 
the robot needs to push the object against one of the four boundaries, pivot to flip and finally grasp it from the top.
However, successful execution of these actions is challenging as even minor errors can have significant negative consequences. For instance, a slight misalignment or inappropriate force can cause the object to tumble unpredictably during flipping. 
Manually designing primitives for these intricate actions would be a difficult and error-prone task, further highlighting the need for a more adaptive approach to addressing these challenges.

Agarwal et al.~\cite{agarwal2019model} proposed to first learn a set of primitives and then a combination to address the task. However, they only show results from simulation. They assume that the object pose and velocity are known, which is strong limitation for real-world applications. Instead, we use visual input to train our approach in simulation through curriculum learning~\cite{portelas2020automatic} and apply Automatic Domain Randomization~\cite{akkaya2019solving} to enable zero-shot transfer to the real world, achieving a picking success rate of 98\% even when the object is placed in a random initial location. 
%
%
Our main contributions are thus: 
First, we propose a novel method for solving the occluded grasping task using hierarchical reinforcement learning.
Second, we devise a curriculum learning strategy that allows us to train the low-level agents before progressing to high-level decision-making.
Lastly, we demonstrate our method trained in simulation and achieve zero-shot transfer to a real-world robot experiment.



\section{\uppercase{Related Work}}
\label{sec:related_work}

\subsection{Primitive-based Robotic Manipulation}
Reinforcement learning for robot manipulation poses a significant challenge due to the difficulty of effectively exploring the high-dimensional continuous action space and the complexity of contact-based dynamics.
To address these problems, early works~\cite{zeng2018learning, ren2021fast, yang2021collaborative, xu2021efficient, tang2021learning, dalal2021accelerating} have explored reinforcement learning for manipulation using pre-defined primitives. Instead of exploring the high-dimensional continuous action space, their policies learn to estimate the appropriate primitive and its parameters, such as starting pose, moving distance, or rotation angle.
Recent work~\cite{nasiriany2022augmenting} applies hierarchical reinforcement learning~\cite{kulkarni2016hierarchical} for separating the primitive and the estimation of its parameters to improve performance. They use an \textit{atomic primitive} that directly controls the end-effector pose to fill the missing gaps that cannot be fulfilled by the available primitives.

Although these works demonstrate significant results in using primitives for manipulation tasks, they all rely heavily on manually designed primitives, which often requires human expertise and takes a significant amount of time and effort. In contrast, we learn the behavior of an 
extrinsic dexterity primitive for flipping flat objects by hierarchical reinforcement learning without designing it manually.

\subsection{Extrinsic Dexterity for Manipulation}
In many practical applications, robots are equipped with parallel jaw grippers that are simple, but limited in dexterity, and thus often insufficient for accomplishing more complex tasks. \textit{Extrinsic dexterity}~\cite{dafle2014extrinsic} is one strategy to mitigate this issue by exploiting external resources such as gravity, external contacts, or dynamic motions for assisting manipulation. 
Early works~\cite{eppner2015exploitation, eppner2015planning, bimbo2019exploiting} have proposed exploiting constraints imposed by the environment and manually designed controllers to grasp objects by sliding and pushing against a wall or sliding to the edge of a table. These approaches are, however, limited to controlled conditions and known environments. 
%
Current non-prehensile object re-arrangement approaches~\cite{zhou2023learning, kimpre, zhang2023learning, cheng2023enhancing} utilize reinforcement learning to overcome the inherent unpredictability of physical interactions. 
However, they often suffer from prohibitively high interaction sample complexity and are constrained by their reliance on precise knowledge of the object pose, or the necessity for specialized policies for diverse objects.
%
Recent works~\cite{liang2021learning, sun2020learning} learn a policy to grasp flat objects based on visual information. However, they rely on simple visual servoing to initiate grasping, assume the object position is given, or need a specific gripper design. In contrast, we use a standard parallel jaw gripper, based on visual information and without a given object position or grasp pose.

Zhou and Held~\cite{zhou2022ungraspable} are closely related to our work and also address grasping objects placed in unfavorable configurations through reinforcement learning. 
Their method focuses on learning a controller that is able to flip an object and acquire an initially occluded grasping configuration.
However, the approach presented has several limitations: the target object needs to be placed very close to a wall, a target grasp configuration needs to be available, and the object pose has to be tracked through the interaction. 
In contrast, we combine different primitives to overcome the constraint on object and wall proximity and demonstrate our approach is able to efficiently solve the problem without access to a target grasp or object pose estimate.


\section{\uppercase{Method}}

\begin{figure*}[t!]
    \centering
    \includegraphics[width = 0.95\linewidth]{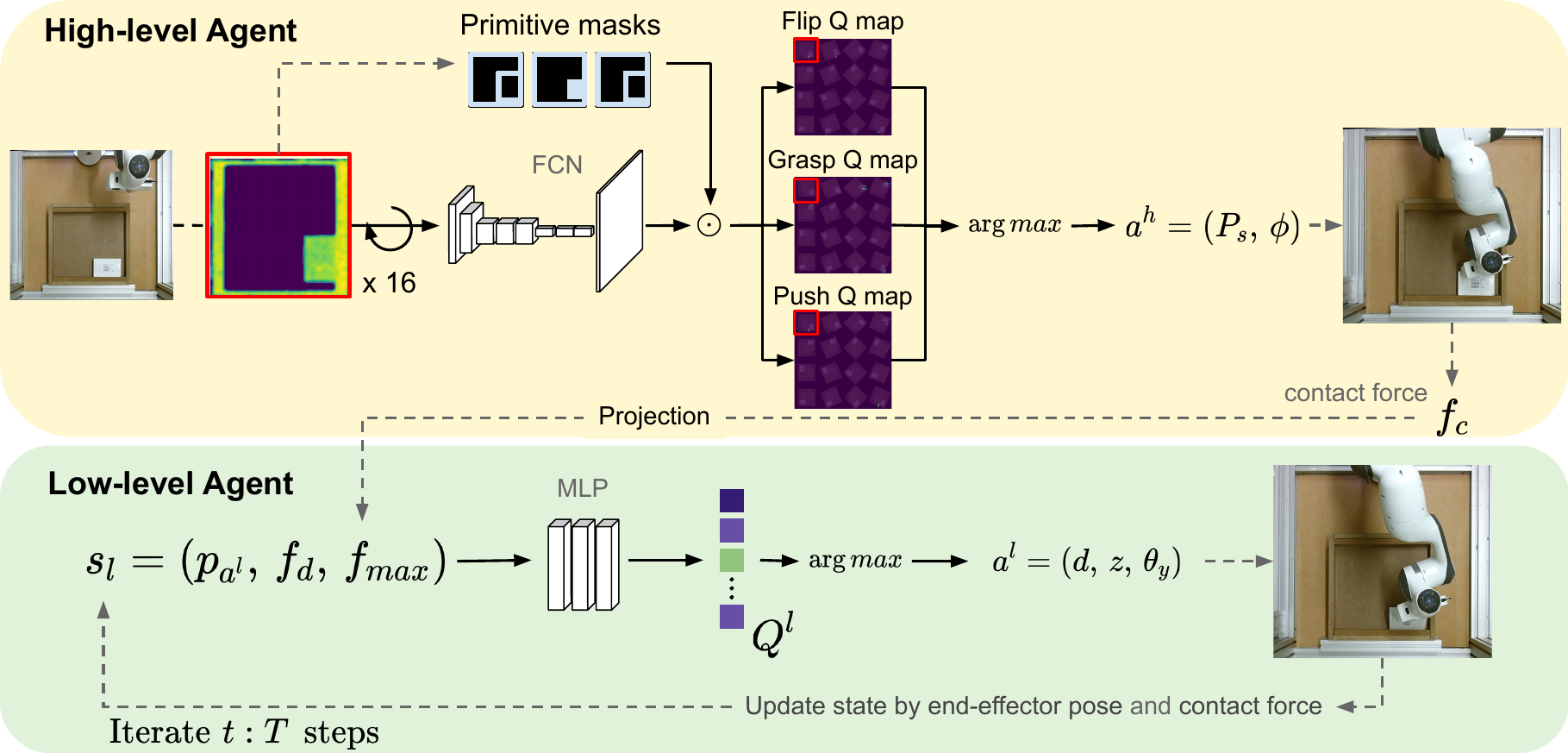}
    \vspace{-0.2cm}
    \caption{\textbf{Overview}. Our ED-PMP method aims to break down complex tasks into sub-tasks and reduces the need for manual primitive design. It comprises high-level and low-level agents.
    \textbf{High-Level Agent} (top): The high-level agent takes a height map as input to a DQN, implemented using an FCN model. It then outputs pixel-wise maps of Q values, where each pixel corresponds to a starting pose and a primitive. 
    \textbf{Low-Level Agent} (down): The low-level agent combines the current end-effector pose and contact force as the state of a DQN model. It iteratively estimates a series of actions to accomplish the sub-task within a designated number of iterations, denoted as $T$.
    }
    \label{fig: method}
    \vspace{-0.5cm}
\end{figure*}

We address the \textit{occluded grasping} task by employing a variation of hierarchical Deep Q Networks (DQN)~\cite{kulkarni2016hierarchical}. 
In our approach (see Fig.~\ref{fig: method}), a high-level agent is responsible for selecting a sequence of pose-parametrized manipulation \textit{primitives},
each of which is in turn assigned to a low-level agent responsible for selecting appropriate primitive-specific actions. 
%
%
The goal of the high-level agent is to learn a policy that maps a sensor observation in the form of depth data to an appropriate pose-parametrized manipulation primitive. 
The low-level manipulation primitive constitutes a feedback controller that is learned through interaction.

We use three manipulation primitives: 
a \textit{push primitive} that achieves in-plane object motion; 
a \textit{flip primitive} that uses contact with the environment to pivot an object, and 
a \textit{grasp primitive} that picks directly graspable objects. 
These three primitives can be combined to solve the \textit{occluded grasping} task and demonstrate extrinsic dexterity manipulation. For example, the high-level agent may first decide to use the \textit{push primitive} to push the object to the wall, then use the \textit{flip primitive} to pivot the object, and finally use the \textit{grasp primitive} to grasp it.
In this paper, we employ a low-level DQN agent to learn the complex \textit{flip primitive}, while we design the other two primitives manually.

\subsection{Manipulation with Parameterized Primitives}
\textbf{Observation and action spaces.} 
The observation space of the high-level agent is a depth map of the workspace.
The action space is composed of a pixel coordinate that corresponds to samples of the state-action-value function Q. Each pixel represents a starting pose $(x, y, \theta_i)$ and a primitive id $\phi$, where 
$(x, y)$ encodes the $(x, y)$-th pixel of the depth image; 
$\theta_i = 2\pi i / K$ encodes the $i$-th discrete end-effector rotation of $K$ possible directions around the $Z$-axis ($K$=16 in this paper); while
$\phi$ is a categorical choice between a group of primitives $\Phi$, with $\|\Phi\|=3$.

The policy estimates a separate Q map for each possible primitive choice. The optimal action then corresponds to the pixel with the maximum Q-value, potentially within a masked region of interest as we discuss later on in this section. 
Based on the action, the robot moves to the starting pose and waits to execute the selected primitive. The height of the starting pose is decided by the height in the corresponding depth map (x, y)-th pixel.

\textbf{Reward}. 
We train the high-level agent using a sparse reward for each primitive. 
Successful actions with the \textit{flip primitive} and the \textit{grasp primitive} are rewarded by $r^{H_f}_t=1$ and $r^{H_g}_t=1$ respectfully. 
The reward of successful \textit{push primitive} actions is set to a value of $r^{H_p}_t=0.2$ on success and $r^{H_p}_t=0.1$ on a change in the workspace configuration. The values are kept lower than the rewards for flipping and grasping to discourage the agent from spending the entire episode just pushing the object. 
Executing a \textit{flip primitive} is considered as success if the object is flipped up after applying the primitive. 
The \textit{grasp primitive} succeeds when the target object is grasped at a certain position. 
Finally, the \textit{push primitive} is successful if the object is pushed to a configuration near one of the four workspace walls. 
Because the reward function of the high-level agent is decided by the results of the primitive execution, we train the high-level agent together with the low-level agent.

\textbf{Primitive masks}. 
We train the high-level agent with $\varepsilon$-greedy exploration and adopt the masking approach from Ren et al.~\cite{ren2021fast} to improve learning efficiency.
As the initial policy is likely to cause no change in the workspace state (e.g., if the robot does not touch any object), we encourage the agent to explore more widely by uniformly sampling an action that corresponds to one of the top ten Q values. 
To further improve learning efficiency, we apply primitive masks to reduce the region that the agent needs to explore.
The idea is that the agent only has to explore pixels near the boundary of the target object. 
Thus, we calculate the mask from the height map: first, we calculate the grasp mask by checking for pixels above a threshold $M_h$; second, we devise the flip and push masks by diffusing an area around the grasp mask. 
With these optimizations, the high-level agent is ready to choose and apply a low-level action primitive.

\textbf{Model architecture}.
We use a Fully Convolutional Network (FCN)~\cite{long2015fully} as our high-level agent model, inspired by Ren et al.~\cite{ren2021fast}. 
As indicated in Fig.~\ref{fig: method}, we pre-process the input depth image to obtain the state space for our high-level agent. 
First, we transform the depth image to a robot-centric coordinate frame and project it to a height map relative to the table surface. 
The height map is then rotated K times for the angles $\theta_i = 2\pi i / K$ and concatenated as batches of tensors that we pass through the FCN.
The FCN outputs three Q maps corresponding to the primitive choices for each rotated height map. 
After applying primitive masks to the corresponding Q maps, we sample the action by $\varepsilon$-greedy. Finally, the robot moves to the starting pose and waits to execute the selected primitive.

\subsection{Learning Behavior for A Contact-rich Primitive}
\label{subfig:low-level}

\textbf{Overview.}
A key distinguishing feature of our approach is that we do not rely solely on expert-devised behavior primitives, but rather learn low-level action policies in conjunction with the high-level agent from the previous section. 
In this paper we train only one such low-level agent --- for the flipping primitive --- though extensions to multiple learnable low-level agents are in principle possible. 

\begin{figure}[t!]
    \centering
    \includegraphics[width=\linewidth]{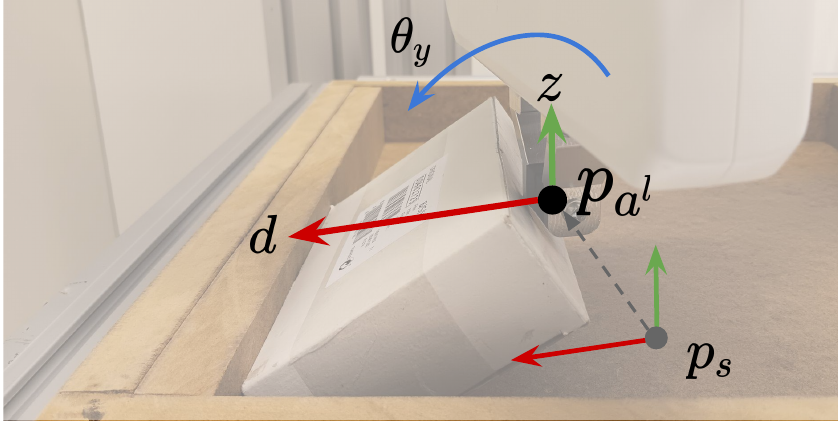}
    \vspace{-0.7cm}
    \caption{
    The end-effector displacement $(d, z, \theta_y)$ corresponds to action space of the low-level agent, where $P_s$ is the starting pose and $p_{a^l}$ is the current end-effector task-space pose.
    }
    \label{fig: low-level-obs-act}
    \vspace{-0.4cm}
\end{figure}

\textbf{Observation and action spaces.}
%
The action space for the low-level agent is discrete. Each action corresponds to a specific end-effector displacement $(d, z, \theta_y)$, shown in Fig.~\ref{fig: low-level-obs-act}. In this context:
$d \in \{0, a_d\}$ encodes the forward distance,
$z \in \{0, a_z\}$ encodes the vertical movement, and
$\theta_y \in \{-r_y, 0, r_y\}$ encodes the angular deviation along the $Y$-axis relative to the primitive's initial pose.
The observation space of the low-level agent is a combination of the end-effector pose and the contact force.
We formulate the state as $s_l = (p_{a^l}, f_d, f_{max})$, where $p_{a^l} \in \mathbb{R}^3$ is the current end-effector task-space pose; $f_d$ is the contact force along the $d$-axis; and $f_{max}$ is the maximum of the current contact force.
To facilitate efficient training, we develop an observation space designed to remain invariant to the starting pose of the low-level agent, ensuring state consistency across different initial poses.

\textbf{Reward.}
To learn the low-level action policy, we could in a straightforward manner define the reward sparsely on successful flips. 
However, as sparse rewards result in less efficient learning~\cite{hu2020learning}, we choose to instead design a reward function that considers the contact force and end-effector position:
\begin{equation} \label{eq:reward}
    r_\tau = r^{H_f}_t + r^L_\tau
\end{equation}
\begin{equation} \label{eq:reward_l}
    r^L_\tau = 
    \begin{cases}
        \min(\sigma, \frac{z_\tau \sigma}{w}) & \text{, if $f_c>0$} \\
        -1 & \text{, if $f_c>f_{limit}$} \\
        0 & \text{, otherwise}
    \end{cases}
\end{equation}
where $f_c$ is the current contact force, $f_\mathit{limit}$ is the maximum safety contact force, $z_\tau$ is the current end-effector height, and $\sigma$ and $w$ are hyper-parameters that normalize the $z_\tau$ and limit the upper bound of the reward.
Based on this reward function, we encourage the agent not only to flip up the object but also to raise it up with contact and avoid applying too much contact force. We find that without the penalty for applying too much contact force, the robot may trigger emergency stops in the real world, making the transfer of policies learned in simulation more challenging.

\textbf{Initial pose invariant state.} 
To further improve the training efficiency, we ensure the low-level model doesn't need to adapt to various initial poses. We establish the starting pose provided by the high-level agent as the reference base frame for $p_{a^l}$. Consequently, $p_{a^l}$ represents a projection of the end-effector pose with respect to the forward direction, vertical displacement, and rotation along the $Y$-axis, all based on the initial pose frame.

\textbf{Model architecture}.
For more details of our low-level agent, we employ a multilayer perceptron (MLP) as the underlying model. As indicated in Fig.~\ref{fig: method}, we combine $p_{a^l}$, $f_d$, and $f_{max}$ as inputs to the MLP, which in turn outputs Q values corresponding to a specific end-effector displacement to control the robot. 
Once selected for execution, the low-level agent controls the robot iteratively, taking a fixed number of actions within a horizon $T$.

\subsection{Curriculum Learning and Domain Randomization}
We follow a curriculum learning~\cite{portelas2020automatic} approach to train our high-level and low-level agents separately.
Learning the high-level agent is hard when the low-level agent is not competent in its sub-task because the high-level reward is conditioned on whether the low-level task is completed successfully.
%
We train the low-level agent first, devising progressively more complex interaction scenarios and only include the high-level model once the low-level policy can successfully flip objects.

To adapt to the domain shift between simulation and the real world and perform zero-shot sim2real transfer, we employ Automatic Domain Randomization~\cite{akkaya2019solving} during the simulation training. In each episode, we randomly sample the object size, friction, and mass to make our method able to address varied box-shaped objects. To overcome the noise of the height map in the real world, we add Gaussian noise and randomly block a few regions as fake reflections in the simulated depth image.


\begin{figure*}[t!]
    \centering
    \includegraphics[width=0.95\linewidth]{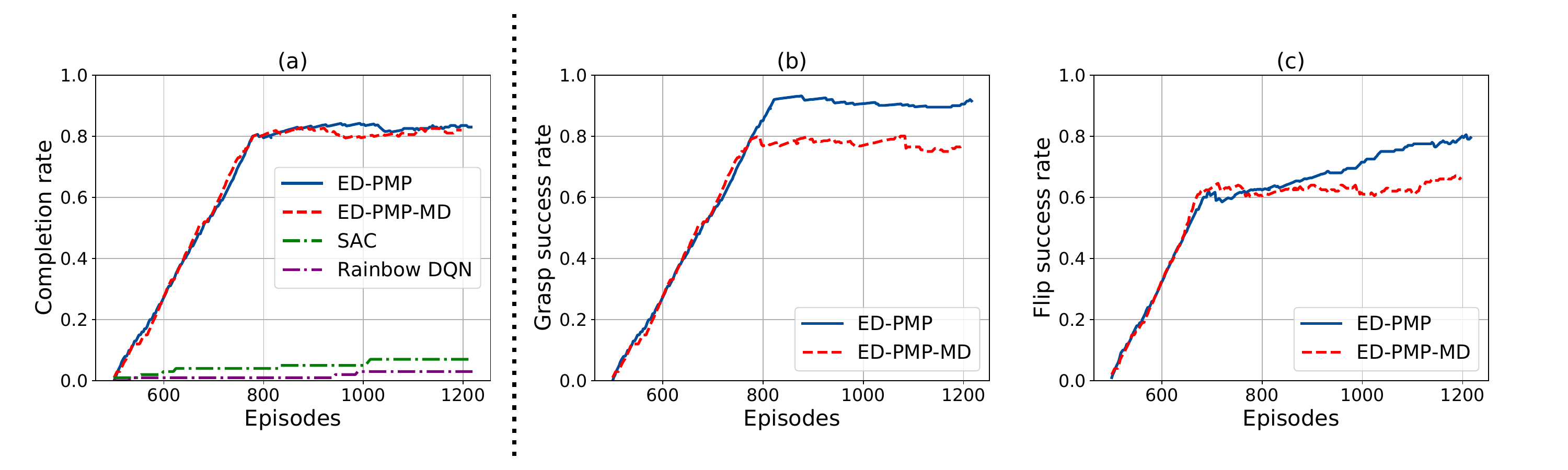}
    \vspace{-0.4cm}
    \caption{
    Testing curve of success rate versus training episodes of the high-level model in simulation. (a) The completion rate for full-task success (successfully picking the object with 10 or fewer primitives). (b) The success rate for the grasp primitive. (c) The success rate for the flip primitive. Success rates of the primitives are computed over the last 100 attempts.}
    \label{fig: exp-sim-curve}
    \vspace{-0.5cm}
\end{figure*}

\section{\uppercase{Experiments}}
\label{sec:experiment}

\subsection{Setup and Evaluation Metrics}

%
To demonstrate our model's ability to learn extrinsic dexterity, we use a Franka Emika Panda arm with a 2-finger gripper that does not open wide enough to grasp the target objects (shown in Table~\ref{table: objects}) from above directly.
%
For the environmental setup, we use the inside of a $44.8 \times 44.8$ (cm) box as the robot's workspace, with the four boundaries serving as potential locations for performing an object pivot. 
An overhead depth image is captured by a Kinect v2 camera and transformed into a height map as input to the high-level model. 
To avoid the robot occluding the workspace, we move the robot to the bottom-left corner before acquiring a depth image.

We use the completion rate as the evaluation metric and test 10 episodes for each object. 
An episode is considered successful if the robot picks up the object within a certain number of primitive actions. 
Similarly to above, the success rate for each primitive is the number of successful actions divided by the total number of attempted actions with that primitive, averaged over the 10 episodes.
This metric does not apply to the baseline algorithms that are not based on manipulation primitives, and thus we omit it when discussing the performance of those algorithms and only report the overall task completion rate.

\subsection{Evaluation in Simulation}

We use Isaac Sim to build a simulation for a variation of the \textit{occluded grasping} task to train our model. 
The hyper-parameters of the low-level model's actions are set to $a_d=0.5$ cm, $a_z=0.5$ cm, and $r_y=2$ degrees. The maximum terminal step $T$ in the low-level model is 35. In the reward function Equation~\ref{eq:reward_l}, we set $\sigma=0.2$ and $w=0.1$.

We train our method in simulation using flat objects of random friction, weight, and size. We randomly place an object in the workspace and evaluate whether the robot can pick it up, applying 10 primitives or less. 
Fig~\ref{fig: exp-sim-curve} shows the learning curve of success rate versus training episodes. We measure the grasp success rate over the last 100 grasp attempts and use the same way to measure the flip success rate and full-task completion rate. In this experiment, we start to test the model after 500 episodes.

We compare the training regime indicated above for our method (ED-PMP) with two baselines: SAC~\cite{haarnoja2018soft} and Rainbow DQN~\cite{hessel2018rainbow}. Additionally, we introduce an ablation --- ED-PMP-MD --- where the learned flip primitive is replaced by a manually designed version.
To adapt the two baselines to our task, we set their action spaces as 6D end-effector movements, with SAC using a continuous version and Rainbow DQN employing a discrete variant. We also increase the maximum episode length from 10 to 40 steps for both baselines.
Regarding ED-PMP-MD, the manually designed flip primitive comprises three stages. First, the gripper moves forward until contact is established. Next, it moves diagonally upward at a 45-degree angle while maintaining the contact force between 8 to 10 N by adjusting the upward and forward movements. Finally, it moves forward while adjusting the upward movement to maintain contact with a force of less than 10 N.

As shown in Fig~\ref{fig: exp-sim-curve}, both SAC and Rainbow DQN struggle to achieve successful object picking with such a limited number of samples.
Comparing our method to ED-PMP-MD, both approaches attain an 80\% task completion rate within 800 episodes. 
However, when analyzing the individual primitive success rates for the grasping (Fig.~\ref{fig: exp-sim-curve} (b)) and flipping (Fig.~\ref{fig: exp-sim-curve} (c)) primitives, we note that in both cases our full method results in higher success rates.
Thus, while in simulation the overall performance of the two methods is comparable, for ED-PMP-MD this comes at the cost of repeated trials: the agent relying on a fixed flipping primitive needs more steps to complete the task. 
We also note that despite the two agents using an \textit{identical grasping primitive} implementation, ED-PMP achieves substantially higher grasping success rates.
We speculate that this is due to the learned flipping primitive manipulating the object into configurations that better afford top-down grasps: a synergy that arises thanks to the simultaneous learning at different hierarchical levels. 

\begin{figure}[t!]
    \centering
    \includegraphics[width=\linewidth]{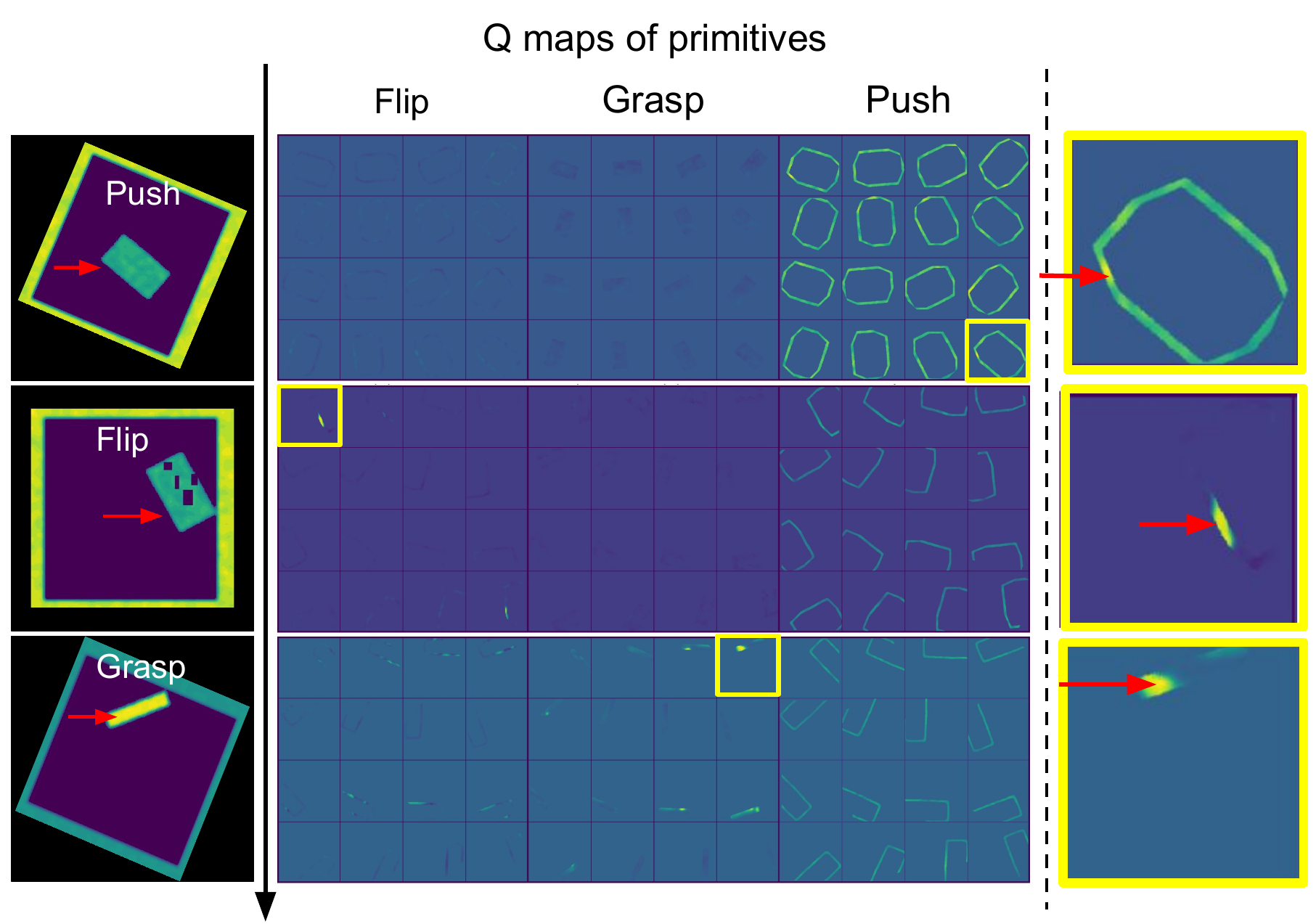}
    \vspace{-0.8cm}
    \caption{An example sequence of picking up a flat object (in simulation). 
    From top to bottom, each row represents a sequence of decisions made by the high-level agent.
    The left column shows the current observation in the form of a height map, while the right column contains the estimated Q maps for each of the three primitives across 16 orientations.
    The maximum Q value and corresponding height map pixel in each step are marked with red arrows. 
    }
    \label{fig:visual-qmaps}
    \vspace{-0.4cm}
\end{figure}

\begin{table*}[t!]
\caption{Real robot experiments with 5 objects in different scenes across 10 episodes}
\label{table:exp_objs}
\setlength{\tabcolsep}{3pt}
\setlength{\tabcolsep}{0.8em} 
\vspace{-0.2cm}
\centering
{\renewcommand{\arraystretch}{1.1}
    \begin{tabular}{p{2.6cm}|c|ccccc|c|c}
        \hline
        Method & Scene & Box-0 (209g) & Box-0 (368g) & Box-1 & Box-2 & Box-3 & \textbf{Average} & \textbf{Mean$\pm$std.dev.} \\
        \hline
        Zhou and Held~\cite{zhou2022ungraspable} & close & 7/10 & 5/10 & 8/10 & 6/10 & 9/10 & 70\% & 7.0$\pm$1.41 \\
        Zhou and Held~\cite{zhou2022ungraspable} & random & 4/10 & 1/10 & 3/10 & 3/10 & 5/10 & 32\% & 3.2$\pm$1.33 \\
        \hline
        ED-PMP-MD & close & 9/10 & 9/10 & 9/10 & 9/10 & 10/10 & 92\% & 9.2$\pm$0.40 \\
        ED-PMP-MD & random & 9/10 & 7/10 & 10/10 & 9/10 & 9/10 & 88\% & 8.8$\pm$0.98 \\
        \hline
        ED-PMP (ours)& close & 9/10 & 9/10 & 10/10 & 10/10 & 10/10 & \textbf{96\%} & 9.6$\pm$0.49 \\
        ED-PMP (ours)& random & 10/10 & 9/10 & 10/10 & 10/10 & 10/10 & \textbf{98\%} & 9.8$\pm$0.40 \\
        \hline
    \end{tabular}
    \vspace{-0.3cm}
}
\end{table*}

To analyze how our agent makes a decision in the current state, we visualize the high-level model's Q maps in Fig~\ref{fig:visual-qmaps} for three consecutive actions executed in simulation. 
As the first action (Fig~\ref{fig:visual-qmaps} top row), the robot moves the object against the right workspace boundary 
by rotating the gripper $-22.5^{\circ}$ around the z-axis and applying the push primitive. 
After pushing (middle row), the robot executes the flip primitive to pivot the object using the right workspace boundary as support. Finally, the robot rotates the gripper $67.5^{\circ}$ around the z-axis and executes a grasp primitive to pick up the object.
When determining the second action (second row in Fig.~\ref{fig:visual-qmaps}), we note that there are very distinct peaks in the Q maps for the flip primitive. The majority of alternative actions for the flip primitive are valued at a uniform low level, indicating that there are only a limited set of primitive parameters that are likely to result in a successful flipping. A similar pattern can be observed for the grasping primitive selected as the third action (third row in the figure).
In contrast, the Q maps for the push primitive at all three decision points consistently exhibit more uniformly bright pixels. This is due to the agent's ability to easily acquire rewards by pushing the object and re-arranging the scene.

\subsection{Real-world Experiments}
\label{sec:exp-real-world}

\begin{table}[t!]
    \caption{Target objects for the real-world experiments}
    \label{table: objects}
    \setlength{\tabcolsep}{0.8em}
    \vspace{-0.2cm}
    \centering
    \begin{tabular}{p{1.4cm}|c|c|c|c} 
        \hline & 
        \includegraphics[width=0.135\linewidth]{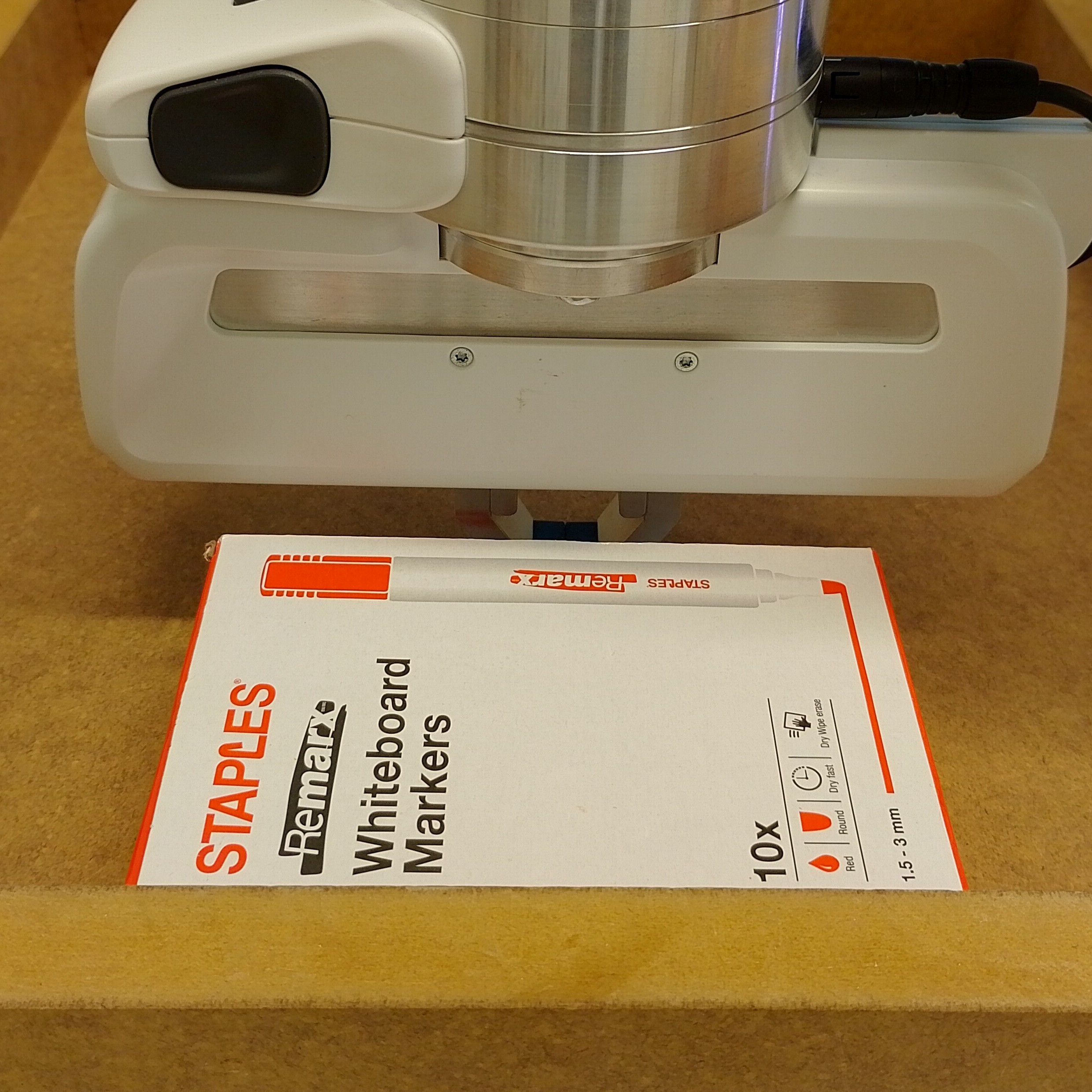} &
        \includegraphics[width=0.135\linewidth]{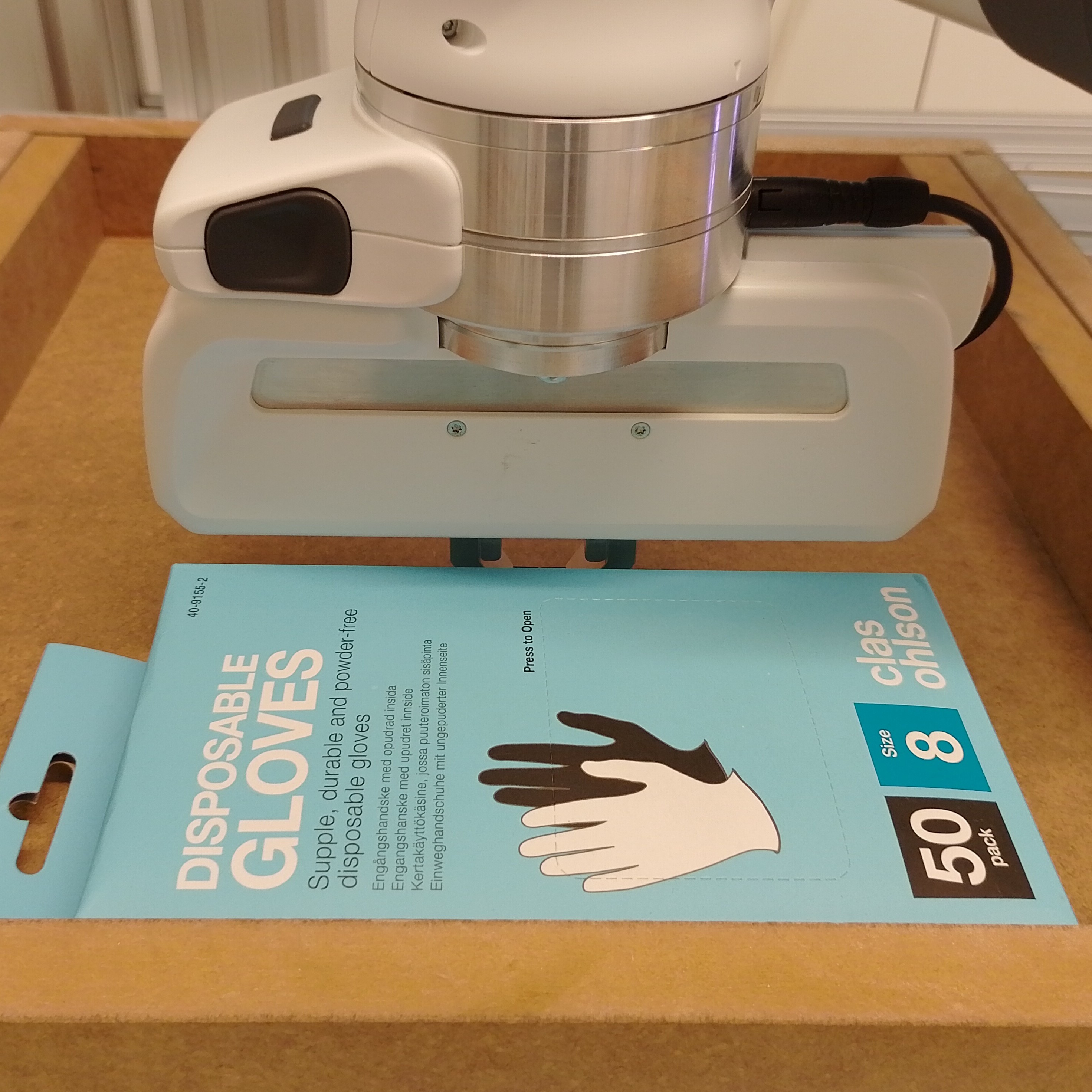} &
        \includegraphics[width=0.135\linewidth]{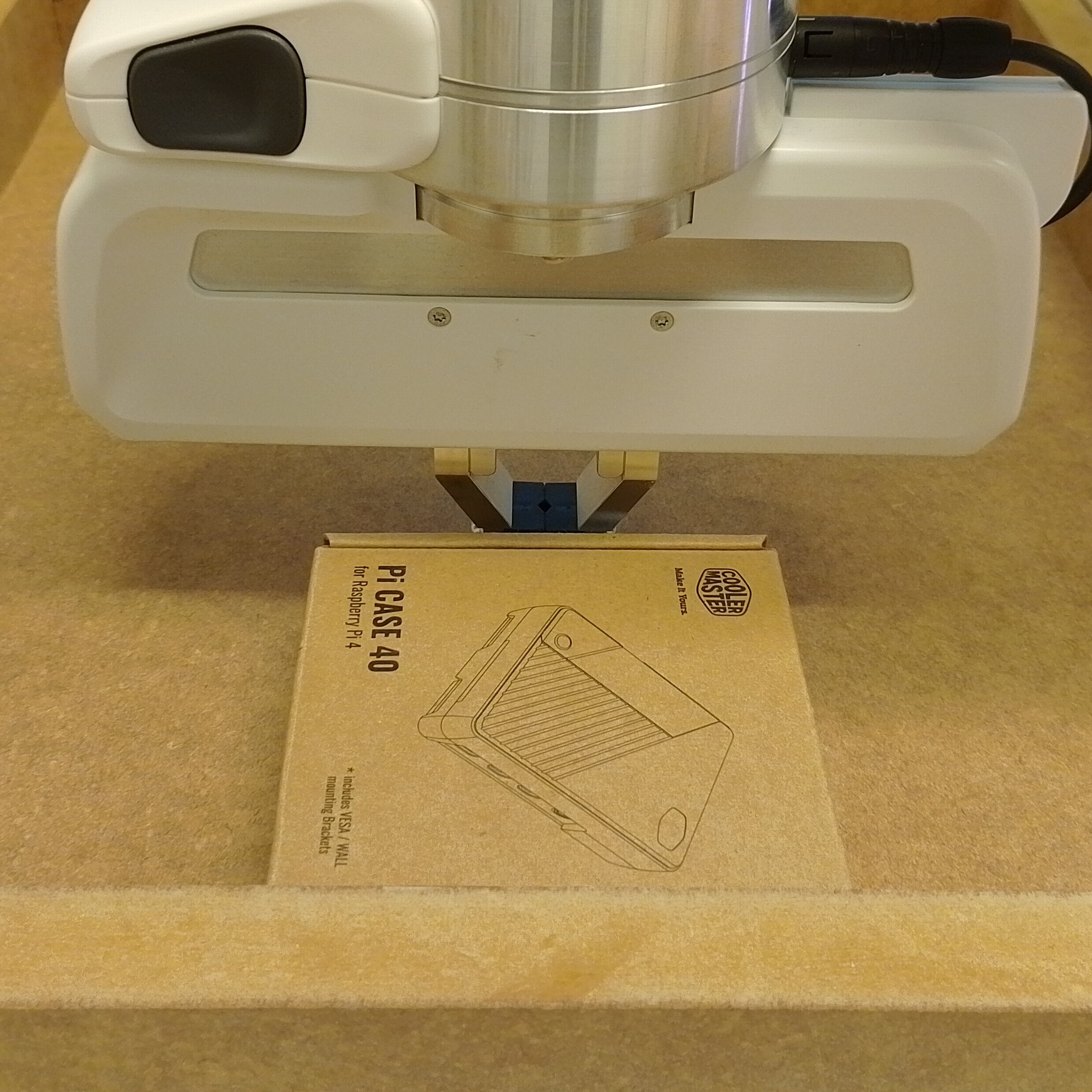} &
        \includegraphics[width=0.135\linewidth]{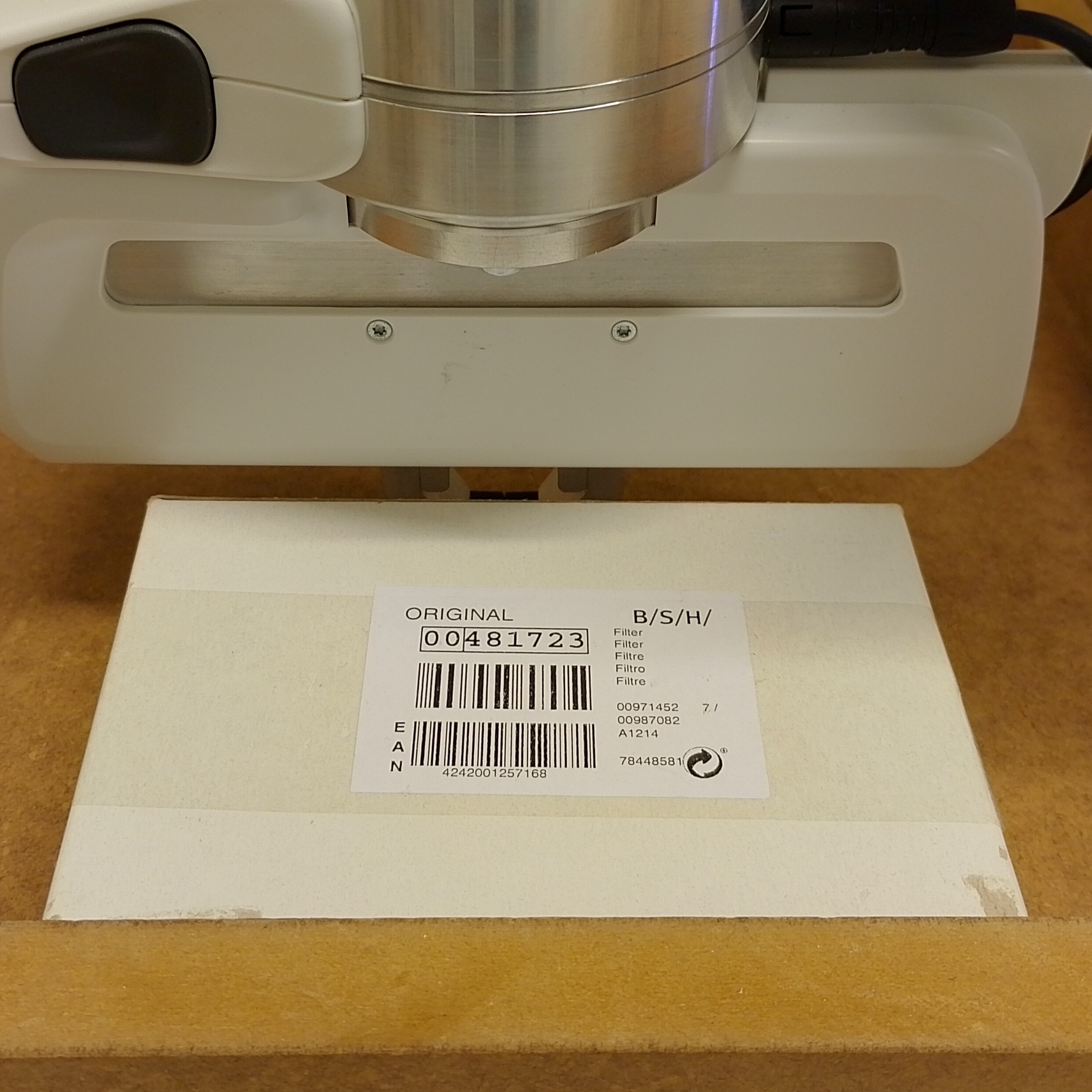} \\
        \hline
        Object-ID & Box-0 & Box-1 & Box-2 & Box-3 \\
        \hline
        Weight (g)  & 209 / 368     & 283   & 58     & 49\\
        \hline
        Length (cm) &  14.2 & 19.9 & 10.4 & 15.1 \\ 
        \hline
        Width (cm)  &  9.1 & 11.2 & 10.2 &10.1 \\ 
        \hline
        High (cm)   &  4.1 & 4.4 & 3.3 & 4.2 \\ 
        \hline
    \end{tabular}
    \vspace{-0.4cm}
\end{table}

To evaluate our method, we test it with zero-shot transfer from simulation to a real-world setup using the box-shaped objects in Table~\ref{table: objects}. 
We use four different boxes (Box-0 to Box-3) but consider two configurations of Box-0 (with and without additional contents to make it heavier), which means we have five object types. 
We consider two setups: \textit{close}, where the object is placed next to one random wall of the box; and \textit{random} where the object is placed randomly near the center of the workspace.
The results are shown in Table~\ref{table:exp_objs}. 

We compare with Zhou and Held~\cite{zhou2022ungraspable} as our baseline. Their method focuses on learning a controller that is able to flip an object and acquire an initially occluded grasping configuration.
Their method achieves completion rates of 70\% and 32\% in the \emph{close} and \emph{random} setup scenes, respectively. It's worth noting that their method encounters significant challenges when the object is not in close proximity to the wall, indicating limitations in its ability to handle such scenarios effectively.
In contrast, our approach with the manually designed flipping primitive (ED-PMP-MD) demonstrates 92\% and 88\% completion rates in the \emph{close} and \emph{random} setup, respectively. Further more, Our method with the learned low-level flipping primitive (ED-PMP) clearly outperforms the manually designed variant. Whether the object is placed close to a wall or randomly within the workspace, our method consistently attains high completion rates, ranging from 96\% to 98\%.
%

We further evaluate our proposed method ED-PMP and ED-PMP-MD considering grasp success rate, flip success rate, and full completion separately. 
As shown in Table~\ref{table:exp_avg}, the method relying on a learned flipping primitive can achieve better performances in terms of completion rate, grasp success rate, and flip success rate in both setups.
Comparing the action efficiency between ED-PMP and ED-PMP-MD, the former requires on average 2.86 and 4.58 primitives to finish the task (in \textit{close} and \textit{random}, respectively) and the latter needs 3.18 and 5.84 primitives. The number of required primitives is significantly less in the \textit{close} setup for both methods since the target object is already placed in a configuration amenable to flipping. 
One interesting observation is that, although the \textit{random} setup is more difficult than \textit{close} one, our method achieves a slightly better completion rate in the \textit{random} setup. We infer the reason is that randomly placing the object in the workspace, as opposed to placing it close to the wall, enables the agent to move the object to a position where it has a higher chance of successfully picking it up.

\begin{table}
    \caption{Average performance in the real-world experiments}
    \label{table:exp_avg}
    \setlength{\tabcolsep}{4pt}
    \vspace{-0.2cm}
    \centering
    \begin{tabular}{p{1.6cm}|c|cccc} 
        \hline
        \multirow{2}{*}{Method} & \multirow{2}{*}{Scene} & \multirow{2}{*}{Completion} & Grasp & Flip & Action \\
               &        &       & Success & Success & Efficiency \\
        \hline
        ED-PMP-MD & close  & 0.92 & 0.921 & 0.622 & 3.18 \\
        ED-PMP-MD & random & 0.88 & 0.937 & 0.408 & 5.84 \\
        \hline
        ED-PMP    & close  & \textbf{0.96} & \textbf{0.967} & \textbf{0.648} & \textbf{2.86} \\
        ED-PMP    & random & \textbf{0.98} & \textbf{0.964} & \textbf{0.602} & \textbf{4.58} \\
        \hline
    \end{tabular}
    \vspace{-0.4cm}
\end{table}

\section{\uppercase{Conclusion}}
\label{sec:conclusion}

In this paper, we propose an approach for solving the occluded grasping task by learning hierarchical control policies that decompose the problem in two steps: choosing a sequence of parametrized manipulation primitives; and learning low-level control policies.
To enhance learning efficiency, we devise a curriculum learning strategy to train the low- and the high-level agents sequentially.
Our method transfers zero-shot to the real world and achieves a 98\% task completion rate with varied box-shaped objects and a wide range of configurations in the real environment. Notably, compared to the state-of-the-art, we do not require the object to be placed near a supporting wall. 
With increasingly complex tasks, designing reward functions for numerous learned parameterized primitives can be challenging.
Thus, a potential future direction is to remove the need for engineering reward functions when training the low-level agent.
	

%



	
	\bibliographystyle{IEEEtran}
	\bibliography{references}
\end{document}